\documentclass[10pt,twocolumn,letterpaper]{article}

\usepackage{iccv}
\usepackage{times}
\usepackage{epsfig}
\usepackage{graphicx}
\usepackage{amsmath}
\usepackage{amssymb}

\usepackage{graphics} 
\usepackage{times} 
\usepackage{amsmath} 
\usepackage{amssymb}  
\usepackage{mathrsfs}
\usepackage{amsfonts}
\usepackage{kantlipsum}
\usepackage{subfigure}
\usepackage{amsmath}

\usepackage{algpseudocode}
\usepackage{algorithm}

\usepackage{graphicx}
\usepackage{float}
\usepackage{ctable}
\usepackage{cuted}
\usepackage{colortbl}
\definecolor{Gray}{gray}{0.9}
\definecolor{White}{gray}{1}
\usepackage{multirow}

\algdef{SE}[DOWHILE]{Do}{doWhile}{\algorithmicdo}[1]{\algorithmicwhile\ #1}%

\usepackage[pagebackref=true,breaklinks=true,letterpaper=true,colorlinks,bookmarks=false]{hyperref}

\iccvfinalcopy 

\def\iccvPaperID{12008} 

\ificcvfinal\fi

\begin{document}

\title{ClothesNet: An Information-Rich 3D Garment Model Repository with Simulated Clothes Environment}

\author{%
Bingyang Zhou$^{1}$, \
Haoyu Zhou$^{2}$\thanks{\ The authors contribute equally to this work.}, \
Tianhai Liang$^{1}$\footnotemark[1],\
Qiaojun Yu$^{3}$,\ 
Siheng Zhao$^{4}$,\
Yuwei Zeng$^{5}$,\
Jun Lv$^{3}$,\
\\Siyuan Luo$^{6}$,\
Qiancai Wang$^{1}$,\
Xinyuan Yu$^{5}$,\
Haonan Chen$^{4}$,\
Cewu Lu$^{3}$,\
Lin Shao$^{5}$\thanks{\ Corresponding author}
\\
$^1$Harbin Institute of Technology, Shenzhen
$^2$Beihang University
$^3$Shanghai Jiao Tong University \\
$^4$Nanjing University
$^5$National University of Singapore
$^6$Xi'an Jiaotong University
\\
}

\maketitle
\ificcvfinal\thispagestyle{empty}\fi

\begin{abstract}
We present ClothesNet: a large-scale dataset of
3D clothes objects with information-rich annotations. Our dataset consists of around 4400 models covering 11 categories annotated with clothes features, boundary lines, and keypoints. ClothesNet can be used to facilitate a variety of computer vision and robot interaction tasks. Using our dataset, we establish benchmark tasks for clothes perception, including classification, boundary line segmentation, and keypoint detection, and develop simulated clothes environments for robotic interaction tasks, including rearranging, folding, hanging, and dressing. 
 We also demonstrate the efficacy of our ClothesNet in real-world experiments. Supplemental materials and dataset are available on our project webpage at~\href{https://sites.google.com/view/clothesnet}{https://sites.google.com/view/clothesnet}.

\end{abstract}

\section{Introduction}
Clothes-related activities, such as folding, laundry, and dressing, play an essential role in our everyday lives. However, achieving autonomous performances of these tasks poses significant challenges in robotics due to the high-dimensional state representation and complex dynamics~\cite{5509439,Miller2012AGA,5980453}. Directly training robots to learn these skills in real-world scenarios can be costly and unsafe. An alternative approach is to develop simulated environments with rich assets where robots can master these skills before transferring to real-world scenes. 

\begin{figure}[t!]
\centering
 \includegraphics[width=0.95\linewidth]{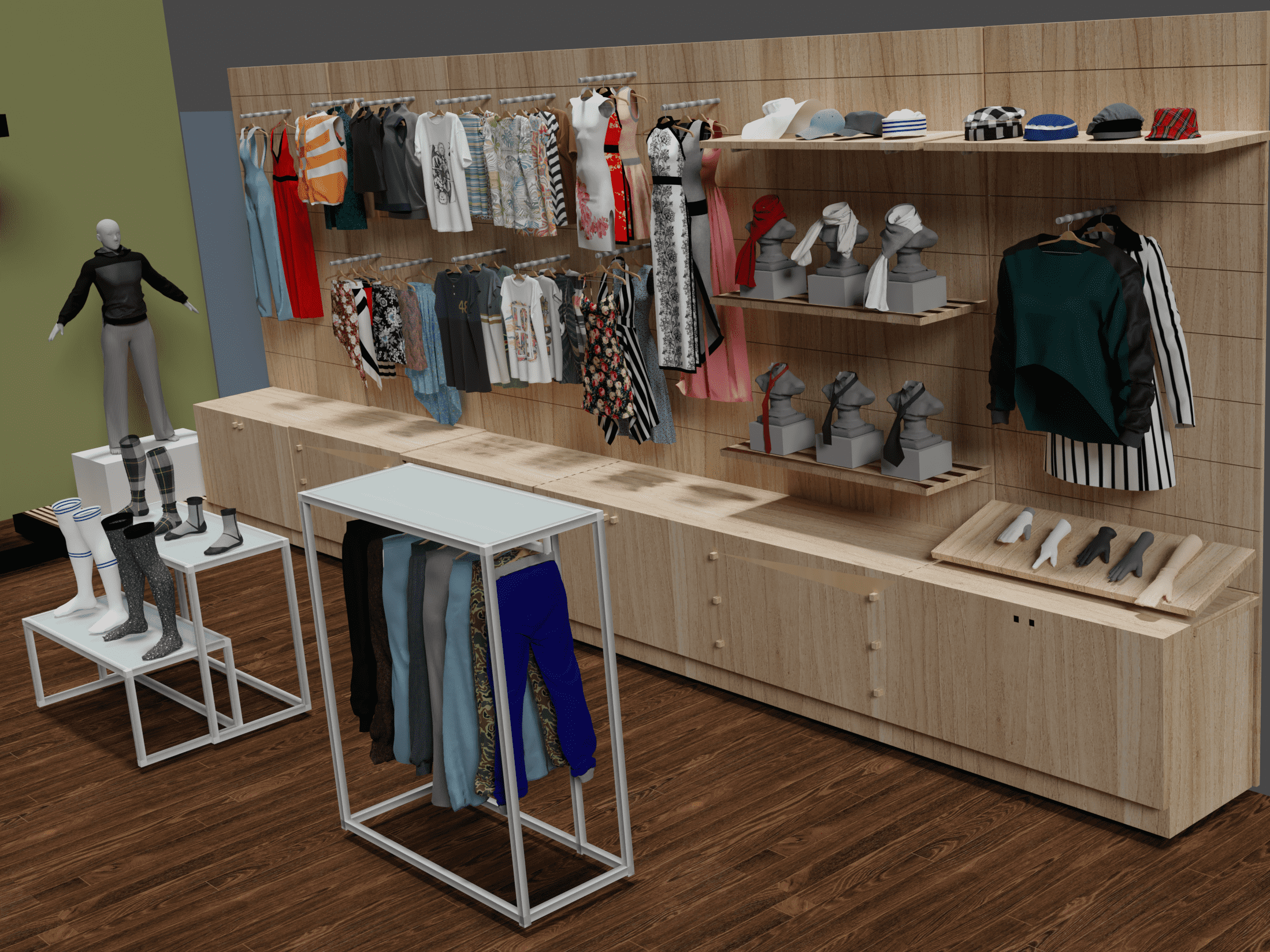}
 \caption{We present ClothesNet consisting of 4400 clothes mesh models covering 11 categories. We annotate ClothesNet with
clothes features, boundary lines, and keypoints. To the
best of our knowledge, it is the first large-scale dataset
with rich annotations for clothes-centric robot vision
and manipulation tasks. We also set up the simulation
environment for robotic manipulation tasks, including the hanging, folding, rearranging, and dressing.}
\label{fig:teaser}
\end{figure}

This learning paradigm often demands large-scale objects with simulation environments for robots to interact with, utilizing data-driven approaches. While there is a growing number of large-scale 3D dataset repositories~\cite{chang2015shapenet,xiang2020sapien,zhou2016thingi10k,koch2019abc}, only a limited number offer 3D clothing models. For instance, Deep Fashion3D~\cite{zhu2020deep} comprises around 2000 3D models reconstructed from real garments across ten categories. SIZER dataset~\cite{tiwari20sizer} includes approximately 2000 scans, including 100 subjects wearing 10 garment classes. However, these scanned 3D models are not suitable for loading into robotic simulations for tasks involving substantial deformation due to data representation or mesh quality. CLOTH3D~\cite{bertiche2020cloth3d} presents a substantial collection of synthetic 3D models with clothing. GarmentNets~\cite{chi2021garmentnets} generates six garment category meshes based on CLOTH3D dataset. Nonetheless, specific categories, such as socks, masks, hats, and ties, are absent from these efforts.


Regarding cloth simulation, differentiable cloth simulation\cite{li2022diffcloth,hu2019difftaichi,qiao2020scalable} demonstrates strong potentials, providing differentiable operations to calculate the gradient information to enhance the cloth dynamics and mitigate the high-dimensional of state and action space.
They provide differentiable operations to calculate the gradient information to enhance the cloth dynamics and mitigate the challenges posed by the high-dimensional of state and action space.  With the availability of these cloth simulations, there are increasing research interests in deformable object understanding. However, they either lack the coupling mechanism with articulated rigid bodies or lead to undesired penetration between cloth-cloth and cloth-articulated rigid body interations. This issue significantly degrades the quality and accuracy of simulations. Addressing this, Yu~\etal~\cite{yu2023diffclothai} introduce ~\emph{DiffClothAI}, a differentiable cloth simulation with intersection-free frictional
contact and the differentiable two-way coupling between cloth and
articulated bodies.

In this paper, we introduce ClothesNet: a large-scale dataset for clothes with rich annotations tailored for robot vision and manipulation tasks. The dataset contains 4400 3D mesh models from 11 coarse categories annotated with clothes features, edge lines, and keypoints. We design clothes robotic manipulation tasks based on the differentiable cloth simulation \emph{DiffClothAI}. We perform benchmark algorithms for clothes classification, edge line segmentations, and keypoint detections. Finally, we demonstrate the usefulness of our dataset by enabling a dual-arm robot to fold clothes in the real-world experiment.  

In summary, we make the following contributions:
\begin{itemize}
\item We create ClothesNet, which contains 4400 3D mesh models from 11 coarse categories, annotated with clothes features, boundary lines, and keypoints. To the best of our knowledge, it is the first large-scale dataset with rich annotations for clothes-centric robot vision and manipulation tasks. 
\item We develop clothes perception tasks and benchmark data-driven methods to demonstrate the usefulness of ClothesNet, including clothes classification, boundary line segmentation, and keypoint detection.
\item We develop clothes manipulation tasks, including folding, hanging, rearranging, and dressing based on a differentiable cloth simulation \emph{DiffClothAI}.
\item We conduct comprehensive experiments both in simulation and real-world setting to demonstrate the efficacy of our ClothesNet.
\end{itemize}


\section{Related Work}
We review related literature, including 3D datasets of clothes, simulation task suits, robotic perception, and manipulation for clothes. We describe how we are different from previous work.
\subsection{3D Garment Datasets}

While there are an increasing number of large-scale 3D dataset repositories such as ShapeNet~\cite{chang2015shapenet}, PARTNET~\cite{mo2019partnet}, SAPIEN~\cite{xiang2020sapien}, Thingi10K~\cite{zhou2016thingi10k}, and ABC~\cite{koch2019abc}, 
only a few datasets consist of 3D models of clothes. BUFF dataset~\cite{Zhang_2017_CVPR} contains high-resolution of 4D scans of clothed humans. It does not provide separate models for body and clothing. MGN~\cite{bhatnagar2019multi} introduces a garment dataset obtained from 3D scans, covering five cloth categories with a few hundred of samples. SIZER dataset~\cite{tiwari20sizer} contains approximately 2000 scans, including 100 subjects wearing 10 garment classes and use ParserNet to extract garment layers from a single mesh. Deep Fashion3D~\cite{zhu2020deep} contains around 2000 3D models reconstructed from real garments under 10 categories and 563 garment instances. 
CLOTH3D~\cite{bertiche2020cloth3d} consists of a large dataset of synthetic 3D humans with clothing. GarmentNets~\cite{chi2021garmentnets} generate six garment categories meshes based on CLOTH3D.
We create ClothesNet Asset, which contains 4400 3D mesh models from 11 categories. A subset of ClothesNet is processed to ensure that these models can be loaded into the differentiable cloth simulation, transforming static 3D clothes mesh into deformable clothes. It provides potential supervision signals to develop data-driven approaches to learn and understand the dynamics between clothes and clothes coupled with articulated rigid bodies.

\subsection{Differentiable Cloth Simulations}
Physically-based cloth simulation is an active research field with diverse applications spanning computer vision, garment design, graphics, and robotics. In recent years, a number of differentiable cloth simulations~\cite{NEURIPS2019_28f0b864,qiao2021differentiable,hu2019difftaichi} has emerged. Although simulating clothing involves complex high-dimensional state and action space, the gradients of the clothes' next state with respect to the current clothes state and action indicate how to improve the action and state such that the clothes' next state moving towards desired/target state. Du~\etal~\cite{du2021_diffpd} design a differentiable soft-body simulator Differentiable Projective Dynamics~(DiffPD) leveraging on Projective Dynamics~\cite{bouaziz2014projective}. Later, Li~\etal extends the \emph{DiffPD} with dry frictional contact to develop a cloth simulation called DiffCloth~\cite{10.1145/3386569.3392396}. However, \emph{DiffPD} and \emph{DiffCloth} do not support the two-way coupling between cloth and rigid bodies. Qiao~\etal~\cite{qiao2020scalable} build a differentiable simulation on top of ARCSim~\cite{10.1145/2366145.2366171} supporting arbitrary meshes and the coupling of deformable object and rigid bodies but not with articulated bodies. Recently, Yu~\etal\cite{yu2023diffclothai} develop a differentiable simulation called \emph{DiffClothAI} with intersection-free frictional contact and the differentiable two-way coupling between cloth and articulated bodies. We design the simulated clothes environment bases on \emph{DiffClothAI}~\cite{yu2023diffclothai} with intersection-free contact modeling and the coupling with articulated rigid bodies, such as the robotic arm and gripper.

\subsection{Clothes Simulation Task Suite}
The field of deformable object simulation environments has witnessed significant progress. SoftGym~\cite{lin2021softgym} presents tasks involving ropes and a rectangular cloth object. Reform~\cite{9561766}, while focusing on linear objects and plastic materials~\cite{huang2021plasticinelab}, lacks
support for thin-shell objects like cloth and garments. DEDO~\cite{antonova2021dynamic} provides a suite encompassing diverse task classes, such as hanging various deformable objects onto rigid hooks and hangers, buttoning with cloth, throwing a rope onto target poles, and putting items onto a mannequin. AssistiveGym~\cite{erickson2020assistivegym} offers a specific assistive dressing task featuring a hospital gown. We develop the clothes simulation environment based on ~\emph{DiffClothAI}~\cite{xinyuan2023} and build environments for clothes folding, hanging, rearranging, and dressing.

\subsection{Robotic Perception for Clothes}
\paragraph{Classification and attribute recognition} To date, deep learning methods have been widely applied for clothes classification and attribute recognition tasks, achieving great success with many applications in fashion field~\cite{hsiao2017learning,10.1007/978-3-642-33712-3_44,9156843}. We annotate clothes with various attributes and class labels. Fig.~\ref{fig::clothlabels} shows a brief overview.
\begin{figure*}[t!]
  \begin{center}
   \includegraphics[width=0.95\linewidth,height=1.0\textwidth]{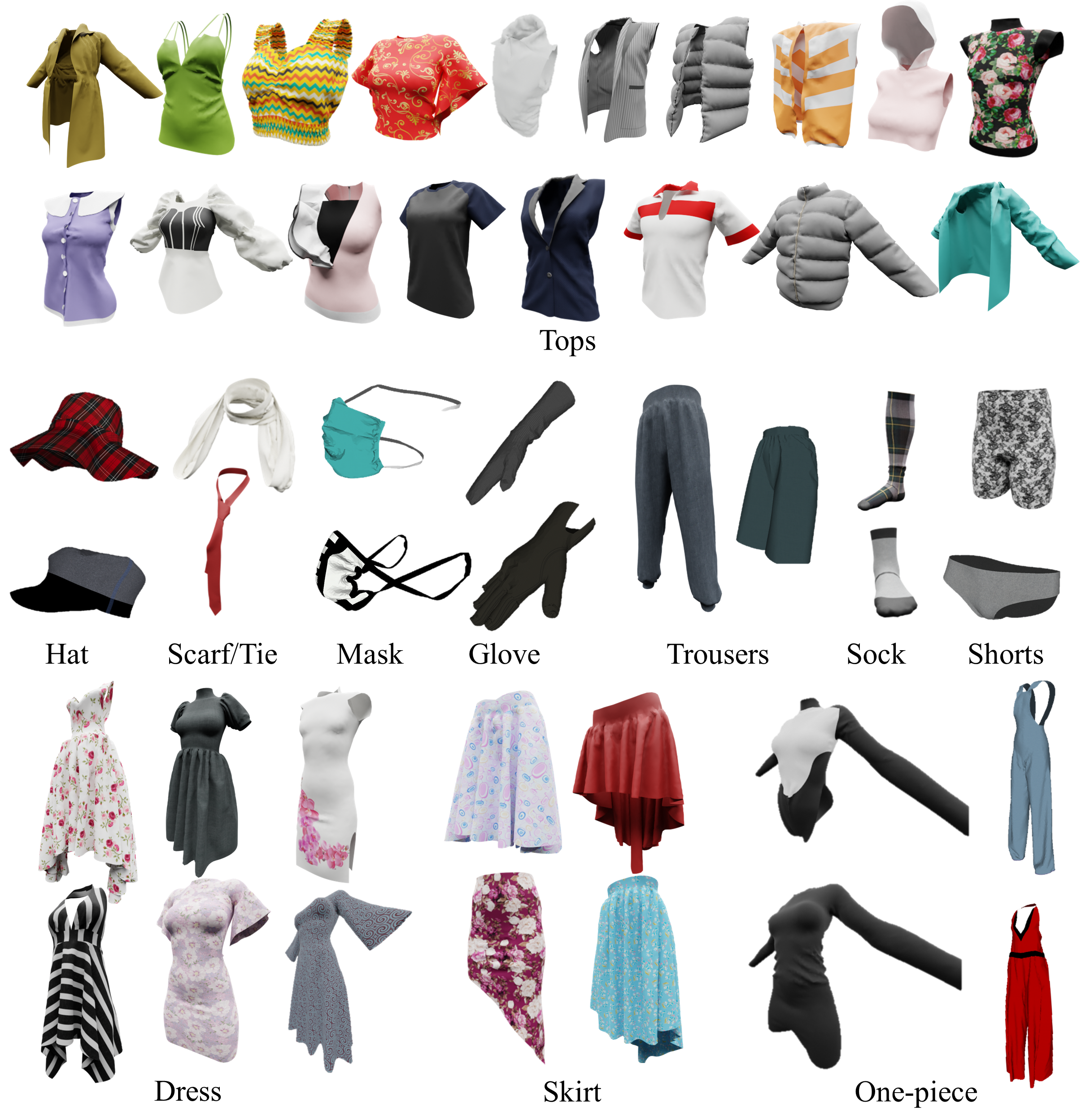}
  \end{center}
      \caption{The overview of our ClothesNet consisting of 4400 clothes mesh models covering 11 categories.}
\label{fig:overview}
\end{figure*}
\paragraph{Segmentation}
Deep Fashion3D~\cite{zhu2020deep} introduces \emph{feature line} annotation which is specially tailored for 3D garments. These feature lines denote the most prominent features of interest, e.g., the open boundaries, the neckline, cuff, waist, etc. that associates with strong priors. The \emph{feature line} has been shown to useful for mesh generations~\cite{zhu2020deep}. Gabas~\etal~\cite{8023660} demonstrates that the physical edges give important clues to determine clothes type and shape as well as to find good grasping points for many manipulation tasks. In ClothesNet, we provide a similar annotation for the boundary line.  We hypothesize
that feature lines are informative for a broad
range cloth insertion tasks, such as human dressing or hanging. Robots need to identify these boundary edges before inserting the garments into human arms/legs or inserting a hanger into the shirts or other clothes.

\paragraph{Self-supervised/Unsupervised Keypoint Detection}
Due to the high dimensional state of clothes, 2D/3D keypoint have been widely adopted as an effective representation for various clothes-related tasks. For \emph{2D keypoint}, Kulkarni~\etal~\cite{kulkarni2019unsupervised} proposed to discover concise keypoints through learning from raw video frames in a fully unsupervised manner. Jakab~\etal~\cite{jakab2018unsupervised} propose a method for learning keypoints detectors for visual objects (such as the eyes and the nose in a face) without any manual supervision.
The use of \emph{3D keypoints} for control is extensively studied in computer vision and robotics~\cite{xue2022useek,shi2021skeleton}. These 3D keypoints are developed and tested for rigid bodies. We provide the results using Skeleton Merger~\cite{shi2021skeleton}.

\paragraph{Clothes Reconstruction and Modeling} So far, various approaches have been proposed to infer the clothes meshes or other parameters from real observations such as images, videos, and point clouds. Zheng~\etal~\cite{zheng20223d} proposed 3D clothing reconstruction method to recover the geometry shape and texture of human clothing from a single 2D image. Sundaresan~\etal~\cite{sundaresan2022diffcloud} presented a clothes modeling approach called \emph{DiffCloud} to estimate clothes meshes from point clouds with differentiable simulation and rendering. Wang~\etal~\cite{Wang:2011:DDE} proposed a piecewise linear elastic material model for cloth, then fit the material model to real cloths by applying controlled forces and measuring the deformation response. Our dataset contains clothes' textures to facilitate the line of works.

\subsection{Robotic Manipulation for Clothes}
There is a rich literature on robotic manipulation of deformable objects. Here we review related manipulation tasks including \emph{folding}, \emph{rerrangment}, and \emph{dressing}. For a broad review, we refer to~\cite{9721534}. Seita~\etal~\cite{seita2021learning} proposed to train robots to learn to rearrange and manipulate deformable objects such as cables, fabrics, and bags with goal-conditioned transporter networks. Corona~\etal~\cite{corona2018active} proposed the search for two grasping points that allow a robot to bring the garment to the target pose. Avigal~\etal~\cite{avigal2022speedfolding} presented a framework for learning efficient bimanual folding policies for garments. Maitin-Shepard~\etal~\cite{5509439} presented a vision-based grasp point detection algorithm to detect the corners of garments relying on only geometric cues that are robust to variation in texture. Hayashi\etal~\cite{7989151,7090661} develop an approach for a bimanual robot to wrap the fabric around a cylinder. Clegg~\etal~\cite{clegg2015animating} described the dressing process as a sequence of primitive actions and developed a set of feedback controllers to chain the primitive actions. Clegg~\etal~\cite{clegg2020learning} presented to use the haptic feedback control and deep reinforcement learning~(DRL) for robot-assisted dressing by simultaneously training human and robot control policies as separate neural networks using physics simulations. Our simulated robotic environments facilitate learning cloth-related manipulation skills.

\section{ClothesNet}
\begin{table*}[t]
\caption{ClothesNetM statistics. We report the class category labels, instances numbers, and the number of vertices. For the sake of convenience in appearance, all data has been directly rounded to the nearest integer.} 
\label{fig::clothlabels}
\centering
\footnotesize
\scalebox{1}{
\begin{tabular}{c|c|ccccc|ccccc}
\hline
\multirow{2}{*}{Category} & \multirow{2}{*}{Instances} & \multicolumn{5}{c|}{Number of points}              & \multicolumn{5}{c}{Number of faces}                  \\
                          &                           & Max       & Min     & Average  & Median  & Std     & Max       & Min     & Average  & Median   & Std      \\ \hline
Trousers                  & 350                     & 22071  & 865 & 7416  & 7124 & 3243 & 43907   & 1662  & 14706  & 14117 & 6464   \\
Dress                     & 408                     & 118344 & 1145 & 10380  & 8199  & 8984  & 235530  & 2121 & 20520   & 16167   & 17890  \\
Mask                      & 49                      & 7046   & 433  & 2585   & 2034 & 1910  & 13456  & 790  & 4836   & 3926  & 3737   \\
UnderPants                & 220                     & 14770  & 546  & 2811    & 2361   & 2070   & 29046  & 962  & 5450    & 4609    & 4098    \\
Hat                       & 109                     & 11319   & 139  & 2787   & 1859   & 2531   & 22510   & 216  & 5477    & 3649    & 5030    \\
Skirt                     & 369                     & 109650 & 1078 & 8960   & 6008  & 9602  & 218205 & 2156  & 17719  & 11857  & 19114  \\
One-piece                   & 146                     & 25787  & 2340 & 8297   & 7725 & 4220  & 51304  & 4514 & 16403  & 15226  & 8435   \\
Glove                     & 96                      & 28053  & 377  & 4966  & 2643 & 5660 & 55650  & 719  & 9837 & 5235  & 11257 \\
Tops                      & 1151                      & 76701   & 171   & 6314    & 5116   & 6538   & 152704  & 304  & 12419   & 10026 
   & 13026   \\
Socks                     & 86                      & 34620  & 877  & 6333   & 4917  & 5100  & 69062  & 1720 & 12582  & 9771   & 10176  \\
Scarf Tie                & 67                      & 21650  & 115  & 5261 
   & 4958   & 4600   & 42632  & 154  & 10231    & 9912   & 9071 
   \\ \hline
Avg                       & 277                      & 42728    & 735   & 6006    & 4813   & 4945  & 84910    & 1393  & 11827   & 9500   & 9834   \\ \hline
\end{tabular}\label{tab:clothnetm}
}
\end{table*}

\begin{figure*}[h]
  \begin{center}
   \includegraphics[width=0.95\linewidth]{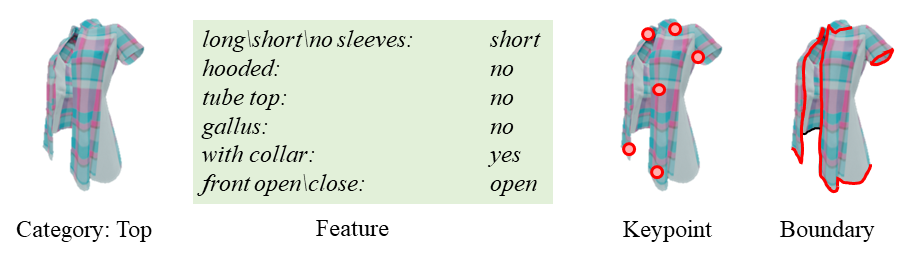}
  \end{center}
      \caption{The illustration of our annotation types.}
\label{fig:anno}
\end{figure*}
We propose a large-scale 3D clothes model dataset that contains around 4400 object models from 11 categories. These categories are tops, dresses, gloves, masks, scarf/ties, skirts, socks, hats, one-piece garments, trousers, and underpants. An overview figure is shown in Fig.~\ref{fig:overview}, indicating our dataset's diversity and high quality. All models in \emph{ClothesNet} are meshes with textures, making them suitable for vision-related tasks.

All models are gathered from~\href{https://www.cgtrader.com/}{CGTrader} and other 3D repositories with licenses to be redistributed for education and research purposes. We apply a series of mesh operations to clean these 3D models and remove duplicate faces and vertices.  We transform quad meshes into triangular meshes. For meshes with too large vertices, we downsample vertices numbers using the quadric edge collapse decimation method~\cite{garland1997surface}. We also perform the weld modifier to mitigate the disconnected components issue, which searches for vertices within a threshold and merges them.

\paragraph{ClothesNetM}
Differentiable cloth simulators exhibit notable advantages for implementing robotics manipulation tasks for Clothes. We publish \emph{ClothesNetM}, a subset of the full ClothesNet dataset containing 3051 models. Each mesh file in \emph{ClothesNetM} satisfies the following three criteria.
\begin{enumerate}
    \item There are no disconnected components: The whole file is a single connected mesh so that the clothing will not split apart when simulating large deformation.
    \item Each garment is a triangle mesh.
    \item No non-manifold edges: Each edge is shared by at most two faces. Non-manifold edges is a common issue in majorities of cloth simulation. Specifically, the differentiable solver cannot construct single dihedral angle constraint for the corresponding vertices~\cite{baraff1998large}. 
\end{enumerate} 
All models in \emph{ClothesNetM} can be directly loaded into \emph{DiffClothAI}. The quality of these models ensures realistic clothes dynamics with large deformation. Table~\ref{tab:clothnetm} summarizes the statistics of \emph{ClothesNetM}.

We annotate the following types of features: clothes category, clothes features, clothes boundary, and clothes keypoint. Fig~\ref{fig:anno} visualizes our annotation types.

\textbf{Categories} We annotate each object's category information. Each mesh file is labeled as one of the categories (tops, dress, gloves, mask, scarf/tie, skirt, socks, hat, one-piece garments, trousers and underpants).

\textbf{Feature Tags} Some categories have diverse design styles. We put feature-rich some categories with attribute tags as follows.
\begin{itemize}
    \item Tops: Whether the top has long sleeves, short sleeves, or no sleeves. Whether it is  hooded, gallus or collared.
    \item Trousers: Whether trousers length is long or short.
    \item Dress: Whether the dress length is long or short. Whether the upper part of the dress exhibits the characteristics described in the tops category.
    \item Skirt: Whether skirt length is long or short.
    \item Socks: Whether socks length is long or medium or short.
\end{itemize}

\textbf{Boundary line}
We annotate the boundary line of a clothes mesh. These boundary lines are the open boundary line such as the neckline, cuff and waist as shown in Fig.~\ref{fig:anno}. They are prominent features for clothes manipulation tasks such as hanging and dressing. They are also important clues to identify clothes type and shape as well as to find good grasping points for many manipulation tasks including folding and rearranging~\cite{8023660,zhu2020deep}. We annotate each mesh's vertices if the vertices belong to the boundary edge by using filter in ~\cite{pymeshlab}.

\textbf{Keypoint} Because of the high dimensional state of clothes, the keypoint has been widely adopted as an effective representation for various clothes-related tasks. We also provide the keypoint annotations for our meshes. We first sample points on the surface of meshes and then run a self-supervised 3D keypoint detection algorithm called Skeleton Merger~\cite{shi2021skeleton} on the point clouds. 

\begin{figure}[h]
\centering
 \includegraphics[width=0.95\linewidth]{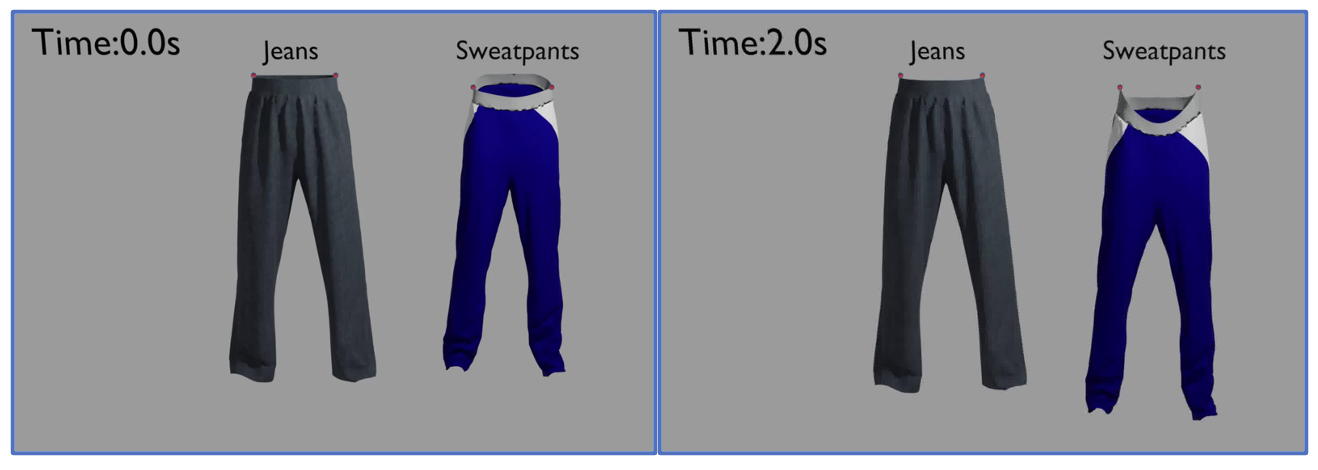}
 \caption{A visualization of different trousers with two grasping points on the waistband under gravity.}
\label{fig:demo}
\end{figure}

\paragraph{Physical Material}
Garments with distinct physical materials yield varied simulation results. For example, jeans are considered larger stiffness than sweatpants. In differentiable simulators like DiffClothAI, sweatpants exhibited greater deformation compared to jeans by setting different physical material parameters (stretch and bending stiffness or other relevant factors). Fig.~\ref{fig:demo} shows the result for visualization reflects these differences with two grasping points on the waistband under gravity. Additionally, our differentiable simulator enables the users to update physical parameters automatically, leveraging the differentiation functioning.

\vspace{-1mm}
\section{Tasks and Benchmarks}
\begin{table*}[thb!]
\caption{Summary of benmark experiment result } \label{Table2DCls}
\resizebox{\textwidth}{!}{
\begin{tabular}{c|ccccccccccc|cc}
\hline
task & Dress & Glove & Hat   & Mask  & One-piece  & Scarf\_Tie & Skirt & Socks & Tops  & Trousers & UnderPants & Class avg & Instance avg \\ \hline
2D classfi(resnet50): Acc   & 0.935 & 1.000 & 0.947 & 0.972 & 0.883   & 0.969      & 0.896 & 1.000 & 0.966 & 0.807    & 0.939      & 0.938     & 0.936         \\ \hline
3D classfi(pointnet):Acc   & 0.688 & 1.000 & 0.938 & 0.900 & 0.528   & 0.875      & 0.602 & 0.773 & 0.890 & 0.783    & 0.869      & 0.804     & 0.796        \\
3D classfi(pointnet++):Acc & 0.683 & 1.000 & 0.938 & 1.000 & 0.681   & 0.938      & 0.878 & 0.955 & 0.962 & 0.683    & 0.911      & 0.875     & 0.870        \\ \hline
3D Segment(pointnet):mIoU   & 0.759 & 0.666 & 0.696 & 0.578 & 0.748   & 0.540      & 0.814 & 0.752 & 0.750 & 0.851    & 0.813      & 0.724     & 0.757        \\
3D Segment(pointnet++):mIoU & 0.792 & 0.762 & 0.822 & 0.723 & 0.794   & 0.731      & 0.830 & 0.814 & 0.813 & 0.834    & 0.851      & 0.797     & 0.809        \\ \hline
\end{tabular}\label{tab:ben}
}
\end{table*}

We benchmark three clothes understanding tasks: Clothes Classification, Boundary line Segmentation, Keypoint Detection. ClothesNet also support a wide variety of robotic interaction tasks, including rearranging, folding, hanging, and long-horizon tasks that require planning such as dressing. 

\subsection{Classification and Segmentation}
\textbf{2D classification} One basic clothes understanding task is identifying the clothes category before performing any advanced robotic vision and manipulation actions. We perform a clothes classification based on the 2D image rendered from 3D meshes, and the detailed process is described as follows.

We render 2455 3D garment meshes into 9820 2D images at four different camera poses through the Blender~\cite{blender}. Eighty percent of the 2D pictures are used as a training set, and the remaining twenty percent of images are used as a test set. Then we selected the commonly used ResNet50 model to train and test the classification task of rendered pictures. The classification results are listed in Table \ref{tab:ben} and the ResNet50 model achieves the classification accuracy of 93.8\%.

\begin{figure*}[h]
  \begin{center}
   \includegraphics[width=0.87\linewidth]{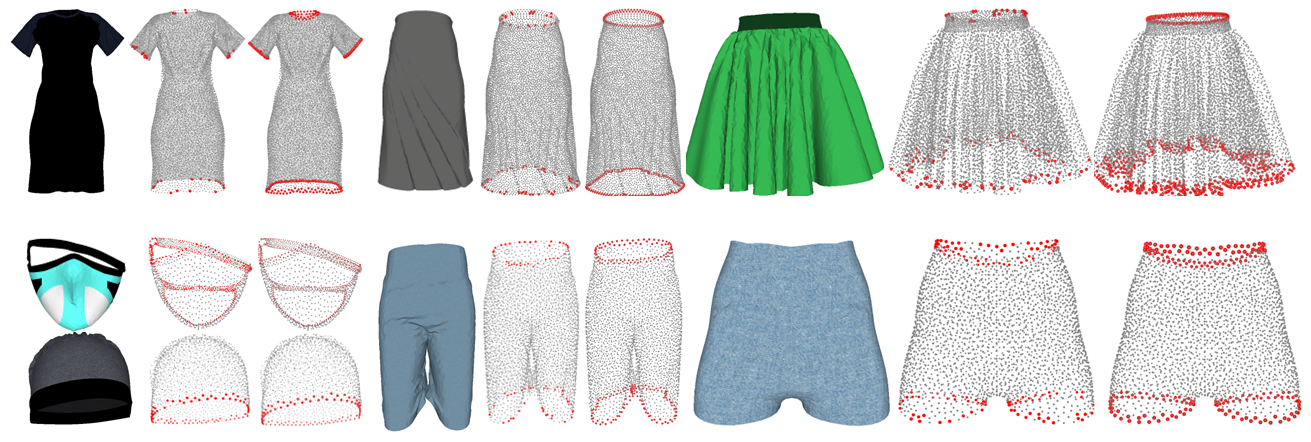}
  \end{center}
      \caption{We visualize the boundary segmentation results of pointnet++ experiment. For each instance pair, the leftmost subfigure shows the clothes. The middle figure is the predicted boundary segmentation result highlighted as red points, and the rightmost subfigure indicates the ground-truth boundary segmentation annotations highlighted as red points.}
\label{fig:boundary}
\end{figure*}

\textbf{3D classification} Leveraging depth sensors, 3D point cloud data is another common modality to represent clothes. We divide the \emph{ClothesNetM} into a training set with 1984 meshes and a test set with 496 meshes. We sample 2048 points on the surface for each mesh using pymeshlab~\cite{pymeshlab}. 

In this experiment, we select Pointnet~\cite{qi2017pointnet} and Pointnet++~\cite{qi2017pointnetplusplus} as the models for the classification task. The experimental result of the classification accuracy is shown in Table \ref{tab:ben}. The PointNet and PointNet++ models achieve the classification accuracy of 80.4\% and 87.5\%, respectively.

\textbf{3D segmentation} For clothes, borders are often the most interesting part of the clothes for various robotic tasks such as hanging or dressing, so we performed a part segmentation experiment on the proposed dataset to identify the borders of clothes. Same as in \emph{3D classification} task, we split \emph{ClothesNetM} into eighty percent training sets and twenty percent test sets.

In the 3D segmentation task, we sample 2048 points on the surface of each clothes mesh. If a point is close to annotated boundary line, then the point's label is one indicating it belongs to the boundary line. Otherwise, the point's label is zero. After gathering the sampled point clouds and ground truth segmentation labels, we feed the processed data into the Pointnet~\cite{qi2017pointnet} and pointnet++~\cite{qi2017pointnetplusplus} model for training. We calculate mIoU for each category and compute the average over all categories as the evaluation metric.
The experimental results are shown in Table \ref{tab:ben}. PointNet experiment achieves the mIoU of 0.724 and PointNet++ experiment achieves the mIoU of 0.797.  We also visualize the predicated quality of pointnet++ experiment as shown in Fig.~\ref{fig:boundary}. Both Table and Figure reflect that our annotated boundary line is reasonable and consistent, and deep network models can learn to identify the boundary lines.

\subsection{Keypoint Detection}
3D keypoint detection is important representation for clothes. We perform keypoint detection on clothesNet and predicted ten keypoints for each cloth, demonstrating that our dataset is suitable for keypoint detection tasks and thus helps for subsequent research.

\begin{figure*}[thb!]
  \begin{center}
   \includegraphics[width=0.87\linewidth]{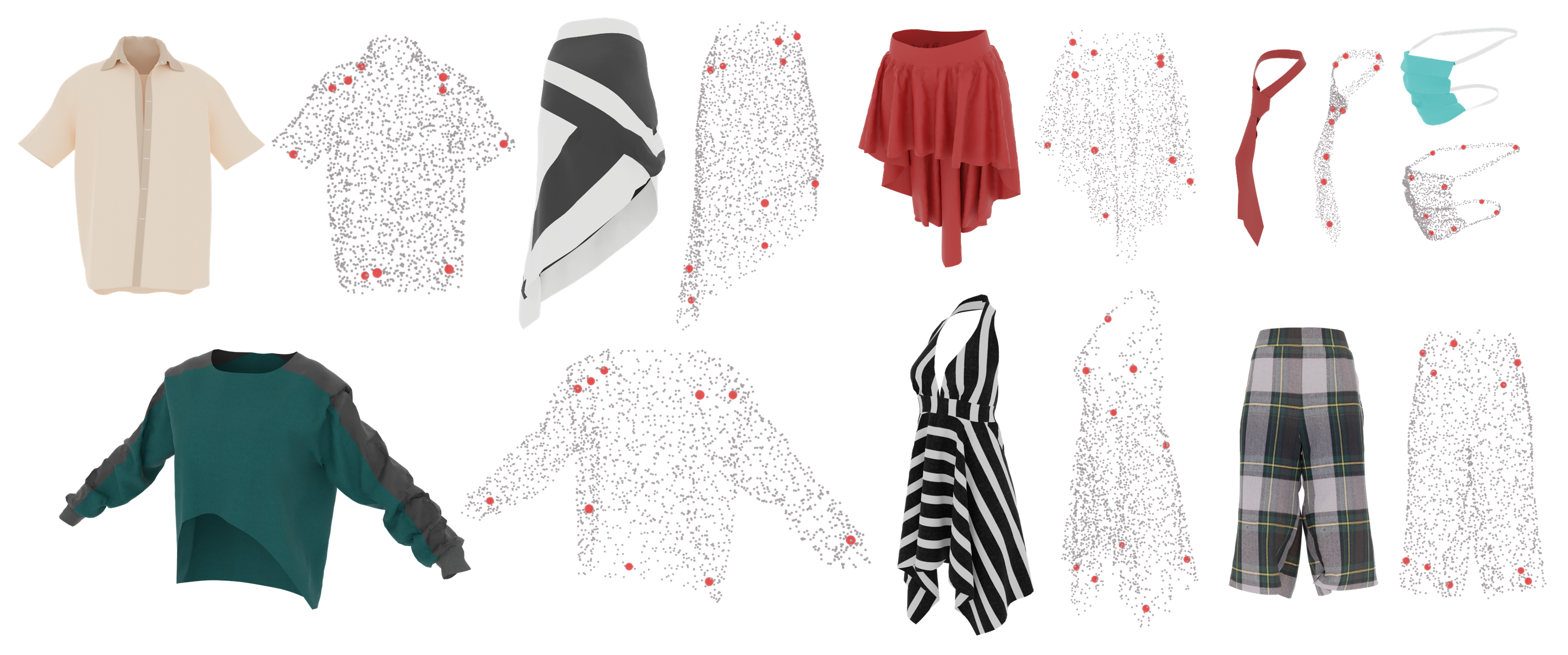}
  \end{center}
      \caption{We visualize the keypoint detection experiment.  For each instance pair, the leftmost subfigure shows the clothes. The right subfigure indicates the learned keypoints highlighted as red points.}
\label{fig:keypoint}
\end{figure*}


We adopt Skeleton Merger~\cite{shi2021skeleton}, a keypoint detector, to train on our clothesNet dataset and make predictions for keypoints. As an unsupervised method, Skeleton Merger shows comparable performance to supervised methods on the KeypointNet dataset, using Pointnet++~\cite{qi2017pointnetplusplus} as a point cloud processing module that can take into account both global information and local details of point clouds. Fig.\ref{fig:keypoint} visualizes the keypoint detection results. It indicates that Skeleton Merger~\cite{shi2021skeleton} learns reasonable keypoints.
\begin{figure}[h]
\centering
\subfigure[\emph{Rearranging}]{
 \begin{minipage}[t]{0.48\linewidth}
    \includegraphics[width=\textwidth]{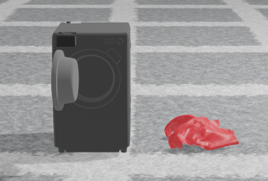}
    \centering
    \end{minipage}}
\subfigure[\emph{Hanging}]{
 \begin{minipage}[t]{0.48\linewidth}
    \centering
    \includegraphics[width=\textwidth]{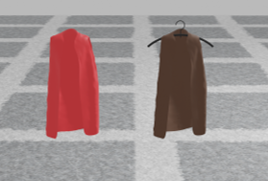}
   \end{minipage}}
\caption{Visualization of the hanging and rearranging tasks. The initial shape of the clothes are highlighted using red color.}
\label{fig:hangrea}
\end{figure}

\subsection{Manipulation Tasks}
The ClothesNet dataset is a comprehensive and large-scale dataset that provides extensive support for various robotic manipulation tasks, including but not limited to grasping, rearranging, folding, hanging, and long-horizon tasks, such as dressing, that require complex planning. The dataset encompasses a wide range of object categories with instance variances, enabling robust and accurate training of robotic systems. The diverse nature of the dataset allows for effective training of robotic systems, facilitating their deployment in real-world scenarios.

\begin{figure}[tbh!]
\centering
 \includegraphics[width=0.8\linewidth]{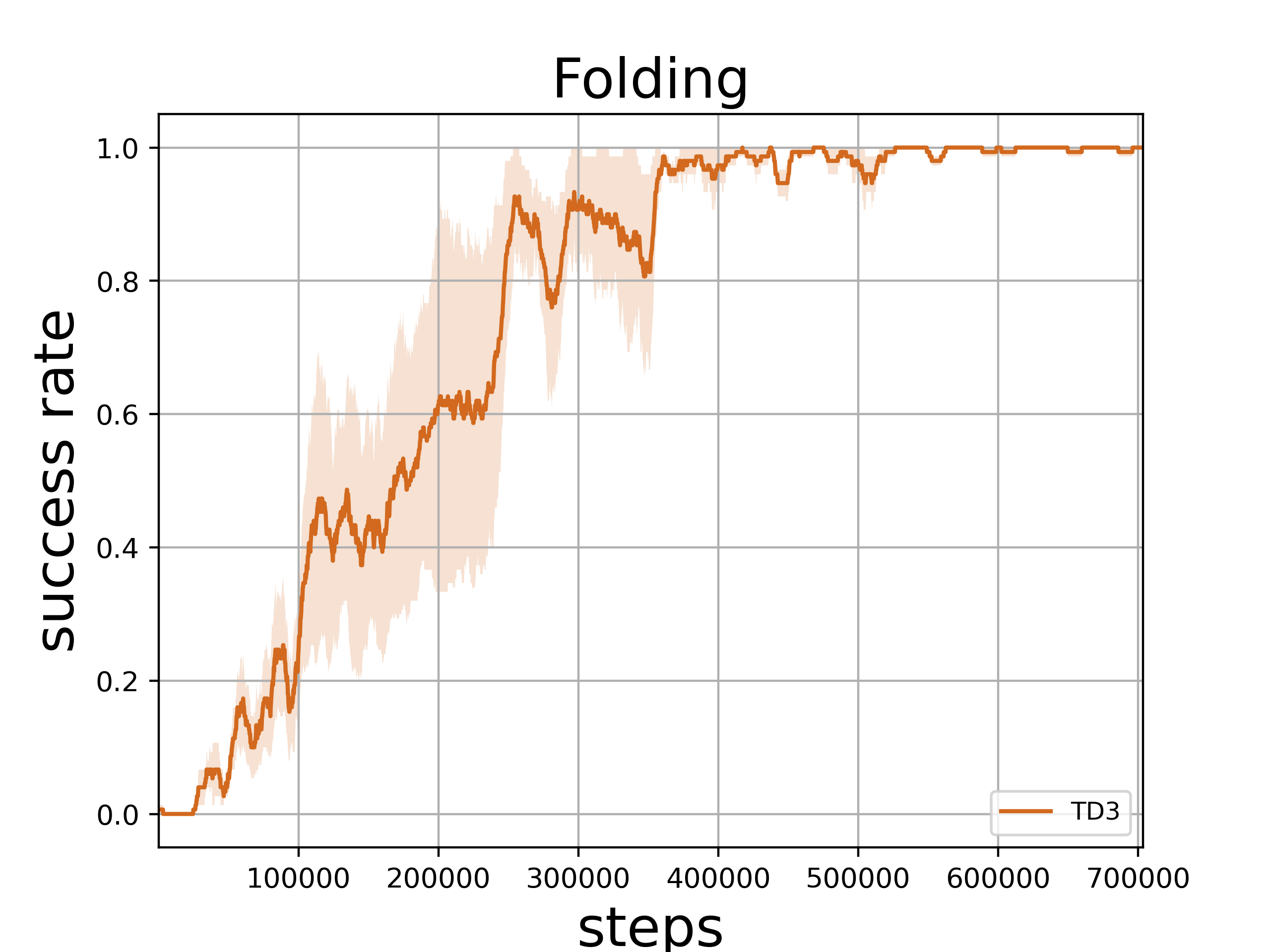}
 \caption{Learning curves. The horizontal axis represents the training steps and the vertical axis indicates the success rates of each task.}
\label{fig:rlcurve}
\end{figure}

\textbf{Folding} 
The goal of the garment folding task is to manipulate specific vertices of the garment mesh to achieve a desired folded configuration. In the reinforcement learning experiment, we use the off-policy algorithm TD3~\cite{fujimoto2018addressing}. To simplify the training process, we use the position and velocity of 20 key vertices as observations, which effectively capture the movement and dynamics of the garment and reduce computational complexity. We selected eight control vertices and manipulated them by specifying their displacement, which served as the action. To encourage the agent to move the garment vertices towards the desired folded configuration, we designed a reward function that is formulated as the negative Euclidean distance between the target and the current position of 20 pre-selected vertices on the garment mesh. The task is considered complete when all 20 pre-selected vertices are within 3 centimeters of their respective targets. The learning curve of the TD3 algorithm is shown in Figure~\ref{fig:rlcurve}.

\textbf{Rearrangement}
For the object rearrangement task, we load the washing machine on the ground, and our goal is to put the garment into the washing machine. The setting is visualized in Fig.~\ref{fig:hangrea}. The task state and action space are the same as in \emph{Folding}. It shows that the rearrangement is difficult than the folding task. The learning curves of the TD3 algorithm is shown on our project website. 


\textbf{Hanging} We load a hanger into the environment, and our goal is to hang the garment on the hanger, as shown in Fig.~\ref{fig:hangrea} by controlling some vertexes of the garment mesh. We choose the position and velocity of each vertex of the garment mesh as the state in the RL algorithm. We select seven vertexes of the garment mesh as the control points and control them by specifying their displacement, which is also the action in our RL algorithm. We report the learning curve of TD3 algorithm in Fig.~\ref{fig:rlcurve}. The hanging task is relatively sensitive to small perturbations, and we observe a performance decrease if we continue to train the agent.


\textbf{Dressing} We load a human model and a ground into the environment, and our goal is to dress the human with the given garment mesh. We put the description and the video in the supplementary material.

\textbf{Differentiable simulation and coupling with articulated rigid bodies}
In addition to the above classic reinforcement learning setting, 
our simulated clothes environments provide the differentiation operations to calculate the gradients information, which enhances the learning process. Our simulation provides the differentiable coupling between clothes and articulated rigid bodies. We report the different setting of these four tasks, including different states and action descriptions, in our supplementary material on the project website.

\subsection{Real-world Experiments}
\begin{figure}[h!]
\centering
 \includegraphics[width=0.95\linewidth]{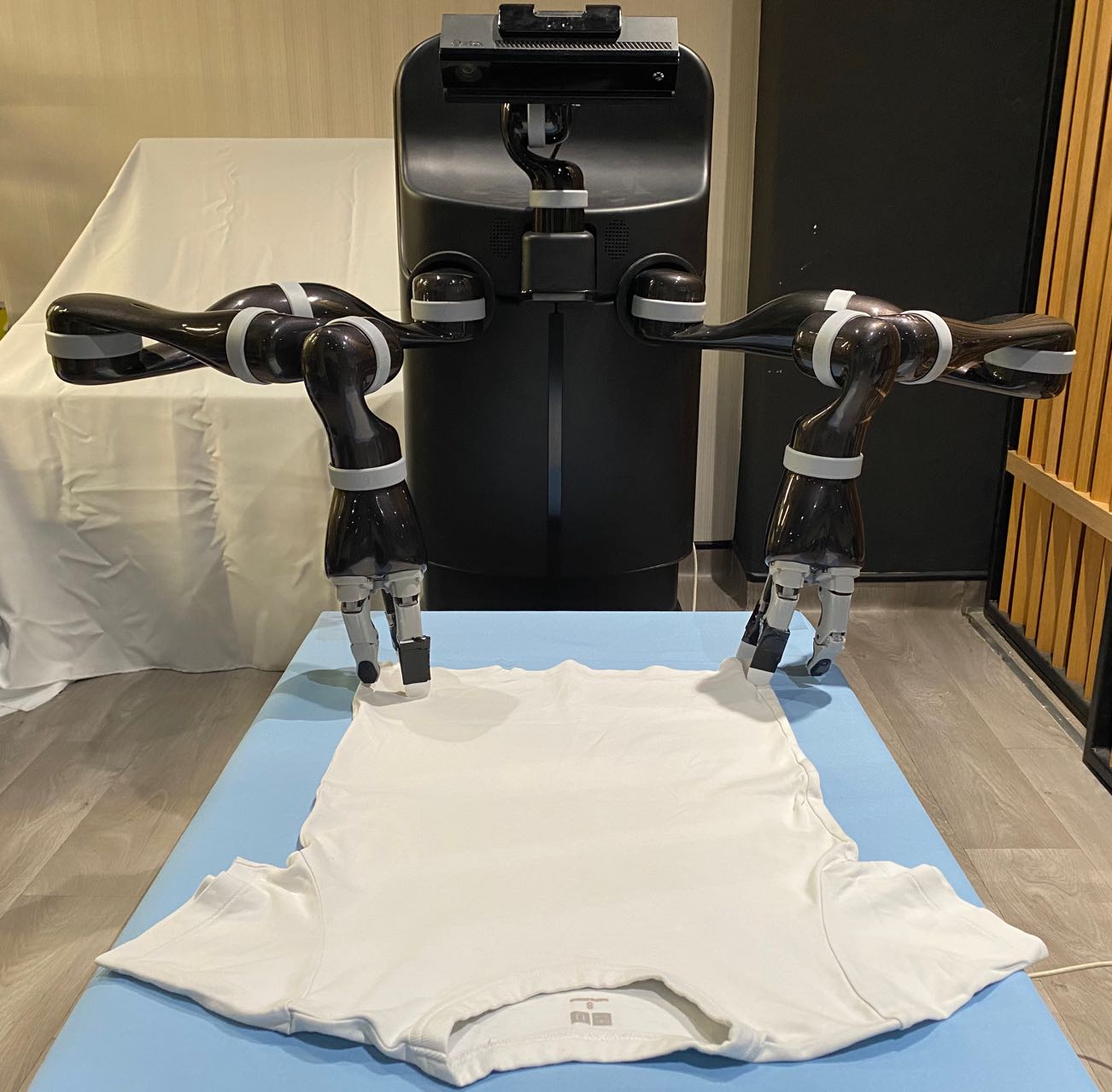}
 \caption{A visualization of our real-world experiment. We collect the point cloud from the RGB-D camera integrated within the MOVO as its head and control the two arms of MOVO to fold the t-shirt.}
\label{fig:realrob}
\end{figure}

\textbf{Folding} 
We perform real-world experiments for cloth manipulation. The experimental setup is shown in Fig.~\ref{fig:realrob}. The t-shirt is put on the table, and the dual-arm Kinova MOVO robot is used to fold the t-shirt. We collect the point cloud from the RGB-D camera integrated within the MOVO as its head. We segment the point cloud using color segmentation to simplify the real-world experiment. More advanced learning-based segmentation is easy to be integrated into the whole framework. We gather the keypoint results by feeding the segmented point cloud into our keypoint detection model. After the keypoints are detected, we control the two grippers to grasp the clothes given the 3D position of the keypoints. We assume the T-shirt is put on the table. So the grasping approach is calculated so that the gripper is grasping along the table. After the gripper is closed, the grippers are controlled to move towards the other side of the t-shirt and then released the gripper. Detailed descriptions and videos are put in the supplementary materials.

\begin{figure}[h!]
\centering
 \includegraphics[width=0.95\linewidth]{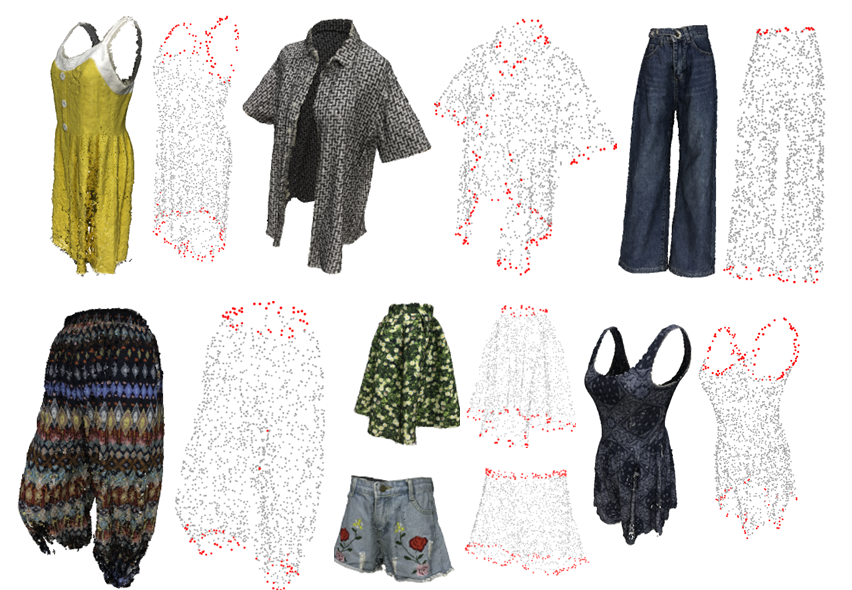}
 \caption{A visualization of our real-world experiment. For each instance pair, the leftmost subfigure shows the raw points cloud on real garments. The right subfigure indicates the predicted boundary segmentation result highlighted as red points}
\label{fig:realseg}
\end{figure}

\textbf{Classification and Segmentation} 
We gather 50 cloth images in real-world from the Internet and 50 raw point clouds on real garments in Deep Fashin3D~\cite{zhu2020deep}, then feed these images/point clouds into our trained models. The accuracy of 2D classification and 3D classification are 82\% and 98\%, respectively. We also visualize the boundary edge segmentation results shown in Fig.~\ref{fig:realseg}, which looks reasonable.

\section{Conclusion}
\label{sec:conclusion}
We introduce ClothesNet, a large-scale dataset of 3D clothing objects annotated with rich information. The dataset contains about 4400 models across 11 categories and has been annotated with clothes features, boundary lines, and keypoints. This dataset can be used for various computer vision and robot interaction tasks. We established benchmark tasks for clothes perception, such as classification, boundary line segmentation, and keypoint detection. We developed simulated clothes environments for robot interaction tasks, such as rearranging, folding, hanging, and dressing. We also conducted real-world experiments to show the effectiveness of ClothesNet.

{\small
\bibliographystyle{ieee_fullname}
\bibliography{egbib}

\begin{thebibliography}{10}\itemsep=-1pt

\bibitem{antonova2021dynamic}
Rika Antonova, Peiyang Shi, Hang Yin, Zehang Weng, and Danica~Kragic Jensfelt.
\newblock Dynamic environments with deformable objects.
\newblock In {\em Thirty-fifth Conference on Neural Information Processing
  Systems Datasets and Benchmarks Track (Round 2)}, 2021.

\bibitem{avigal2022speedfolding}
Yahav Avigal, Lars Berscheid, Tamim Asfour, Torsten Kr{\"o}ger, and Ken
  Goldberg.
\newblock Speedfolding: Learning efficient bimanual folding of garments.
\newblock In {\em 2022 IEEE/RSJ International Conference on Intelligent Robots
  and Systems (IROS)}, pages 1--8. IEEE, 2022.

\bibitem{baraff1998large}
David Baraff and Andrew Witkin.
\newblock Large steps in cloth simulation.
\newblock In {\em Proceedings of the 25th annual conference on Computer
  graphics and interactive techniques}, pages 43--54, 1998.

\bibitem{bertiche2020cloth3d}
Hugo Bertiche, Meysam Madadi, and Sergio Escalera.
\newblock Cloth3d: clothed 3d humans.
\newblock In {\em European Conference on Computer Vision}, pages 344--359.
  Springer, 2020.

\bibitem{bhatnagar2019multi}
Bharat~Lal Bhatnagar, Garvita Tiwari, Christian Theobalt, and Gerard Pons-Moll.
\newblock Multi-garment net: Learning to dress 3d people from images.
\newblock In {\em proceedings of the IEEE/CVF international conference on
  computer vision}, pages 5420--5430, 2019.

\bibitem{bouaziz2014projective}
Sofien Bouaziz, Sebastian Martin, Tiantian Liu, Ladislav Kavan, and Mark Pauly.
\newblock Projective dynamics: Fusing constraint projections for fast
  simulation.
\newblock {\em ACM transactions on graphics (TOG)}, 33(4):1--11, 2014.

\bibitem{chang2015shapenet}
Angel~X Chang, Thomas Funkhouser, Leonidas Guibas, Pat Hanrahan, Qixing Huang,
  Zimo Li, Silvio Savarese, Manolis Savva, Shuran Song, Hao Su, et~al.
\newblock Shapenet: An information-rich 3d model repository.
\newblock {\em arXiv preprint arXiv:1512.03012}, 2015.

\bibitem{10.1007/978-3-642-33712-3_44}
Huizhong Chen, Andrew Gallagher, and Bernd Girod.
\newblock Describing clothing by semantic attributes.
\newblock In Andrew Fitzgibbon, Svetlana Lazebnik, Pietro Perona, Yoichi Sato,
  and Cordelia Schmid, editors, {\em Computer Vision -- ECCV 2012}, pages
  609--623, Berlin, Heidelberg, 2012. Springer Berlin Heidelberg.

\bibitem{chi2021garmentnets}
Cheng Chi and Shuran Song.
\newblock Garmentnets: Category-level pose estimation for garments via
  canonical space shape completion.
\newblock In {\em Proceedings of the IEEE/CVF International Conference on
  Computer Vision}, pages 3324--3333, 2021.

\bibitem{clegg2020learning}
Alexander Clegg, Zackory Erickson, Patrick Grady, Greg Turk, Charles~C Kemp,
  and C~Karen Liu.
\newblock Learning to collaborate from simulation for robot-assisted dressing.
\newblock {\em IEEE Robotics and Automation Letters}, 5(2):2746--2753, 2020.

\bibitem{clegg2015animating}
Alexander Clegg, Jie Tan, Greg Turk, and C~Karen Liu.
\newblock Animating human dressing.
\newblock {\em ACM Transactions on Graphics (TOG)}, 34(4):1--9, 2015.

\bibitem{blender}
Blender~Online Community.
\newblock {\em Blender - a 3D modelling and rendering package}.
\newblock Blender Foundation, Stichting Blender Foundation, Amsterdam, 2018.

\bibitem{corona2018active}
Enric Corona, Guillem Alenya, Antonio Gabas, and Carme Torras.
\newblock Active garment recognition and target grasping point detection using
  deep learning.
\newblock {\em Pattern Recognition}, 74:629--641, 2018.

\bibitem{du2021_diffpd}
Tao Du, Kui Wu, Pingchuan Ma, Sebastien Wah, Andrew Spielberg, Daniela Rus, and
  Wojciech Matusik.
\newblock Diffpd: Differentiable projective dynamics.
\newblock {\em ACM Trans. Graph.}, 41(2), nov 2021.

\bibitem{erickson2020assistivegym}
Zackory Erickson, Vamsee Gangaram, Ariel Kapusta, C.~Karen Liu, and Charles~C.
  Kemp.
\newblock Assistive gym: A physics simulation framework for assistive robotics.
\newblock {\em IEEE International Conference on Robotics and Automation
  (ICRA)}, 2020.

\bibitem{fujimoto2018addressing}
Scott Fujimoto, Herke Van~Hoof, and David Meger.
\newblock Addressing function approximation error in actor-critic methods.
\newblock {\em arXiv preprint arXiv:1802.09477}, 2018.

\bibitem{8023660}
Antonio Gabas and Yasuyo Kita.
\newblock Physical edge detection in clothing items for robotic manipulation.
\newblock In {\em 2017 18th International Conference on Advanced Robotics
  (ICAR)}, pages 524--529, 2017.

\bibitem{garland1997surface}
Michael Garland and Paul~S Heckbert.
\newblock Surface simplification using quadric error metrics.
\newblock In {\em Proceedings of the 24th annual conference on Computer
  graphics and interactive techniques}, pages 209--216, 1997.

\bibitem{7989151}
Naohiro Hayashi, Takashi Suehiro, and Shunsuke Kudoh.
\newblock Planning method for a wrapping-with-fabric task using regrasping.
\newblock In {\em 2017 IEEE International Conference on Robotics and Automation
  (ICRA)}, pages 1285--1290, 2017.

\bibitem{7090661}
Naohiro Hayashi, Tetsuo Tomizawa, Takashi Suehiro, and Shunsuke Kudoh.
\newblock Dual arm robot fabric wrapping operation using target lines.
\newblock In {\em 2014 IEEE International Conference on Robotics and
  Biomimetics (ROBIO 2014)}, pages 2185--2190, 2014.

\bibitem{zhu2020deep}
Zhu Heming, Cao Yu, Jin Hang, Chen Weikai, Du Dong, Wang Zhangye, Cui Shuguang,
  and Han Xiaoguang.
\newblock Deep fashion3d: A dataset and benchmark for 3d garment reconstruction
  from single images.
\newblock In {\em Computer Vision -- ECCV 2020}, pages 512--530. Springer
  International Publishing, 2020.

\bibitem{hsiao2017learning}
Wei-Lin Hsiao and Kristen Grauman.
\newblock Learning the latent" look": Unsupervised discovery of a
  style-coherent embedding from fashion images.
\newblock In {\em Proceedings of the IEEE International Conference on Computer
  Vision}, pages 4203--4212, 2017.

\bibitem{hu2019difftaichi}
Yuanming Hu, Luke Anderson, Tzu-Mao Li, Qi Sun, Nathan Carr, Jonathan
  Ragan-Kelley, and Fr{\'e}do Durand.
\newblock Difftaichi: Differentiable programming for physical simulation.
\newblock {\em ICLR}, 2020.

\bibitem{huang2021plasticinelab}
Zhiao Huang, Yuanming Hu, Tao Du, Siyuan Zhou, Hao Su, Joshua~B Tenenbaum, and
  Chuang Gan.
\newblock Plasticinelab: A soft-body manipulation benchmark with differentiable
  physics.
\newblock {\em arXiv preprint arXiv:2104.03311}, 2021.

\bibitem{jakab2018unsupervised}
Tomas Jakab, Ankush Gupta, Hakan Bilen, and Andrea Vedaldi.
\newblock Unsupervised learning of object landmarks through conditional image
  generation.
\newblock {\em Advances in neural information processing systems}, 31, 2018.

\bibitem{koch2019abc}
Sebastian Koch, Albert Matveev, Zhongshi Jiang, Francis Williams, Alexey
  Artemov, Evgeny Burnaev, Marc Alexa, Denis Zorin, and Daniele Panozzo.
\newblock Abc: A big cad model dataset for geometric deep learning.
\newblock In {\em Proceedings of the IEEE/CVF Conference on Computer Vision and
  Pattern Recognition}, pages 9601--9611, 2019.

\bibitem{kulkarni2019unsupervised}
Tejas~D Kulkarni, Ankush Gupta, Catalin Ionescu, Sebastian Borgeaud, Malcolm
  Reynolds, Andrew Zisserman, and Volodymyr Mnih.
\newblock Unsupervised learning of object keypoints for perception and control.
\newblock {\em Advances in neural information processing systems}, 32, 2019.

\bibitem{9561766}
Rita Laezza, Robert Gieselmann, Florian~T. Pokorny, and Yiannis Karayiannidis.
\newblock Reform: A robot learning sandbox for deformable linear object
  manipulation.
\newblock In {\em 2021 IEEE International Conference on Robotics and Automation
  (ICRA)}, pages 4717--4723, 2021.

\bibitem{li2022diffcloth}
Yifei Li, Tao Du, Kui Wu, Jie Xu, and Wojciech Matusik.
\newblock Diffcloth: Differentiable cloth simulation with dry frictional
  contact.
\newblock {\em ACM Trans. Graph.}, mar 2022.
\newblock Just Accepted.

\bibitem{NEURIPS2019_28f0b864}
Junbang Liang, Ming Lin, and Vladlen Koltun.
\newblock Differentiable cloth simulation for inverse problems.
\newblock In H. Wallach, H. Larochelle, A. Beygelzimer, F. d\textquotesingle
  Alch\'{e}-Buc, E. Fox, and R. Garnett, editors, {\em Advances in Neural
  Information Processing Systems}, volume~32. Curran Associates, Inc., 2019.

\bibitem{lin2021softgym}
Xingyu Lin, Yufei Wang, Jake Olkin, and David Held.
\newblock Softgym: Benchmarking deep reinforcement learning for deformable
  object manipulation.
\newblock In {\em Conference on Robot Learning}, pages 432--448. PMLR, 2021.

\bibitem{10.1145/3386569.3392396}
Micka\"{e}l Ly, Jean Jouve, Laurence Boissieux, and Florence
  Bertails-Descoubes.
\newblock Projective dynamics with dry frictional contact.
\newblock {\em ACM Trans. Graph.}, 39(4), aug 2020.

\bibitem{5509439}
Jeremy Maitin-Shepard, Marco Cusumano-Towner, Jinna Lei, and Pieter Abbeel.
\newblock Cloth grasp point detection based on multiple-view geometric cues
  with application to robotic towel folding.
\newblock In {\em 2010 IEEE International Conference on Robotics and
  Automation}, pages 2308--2315, 2010.

\bibitem{5980453}
Stephen Miller, Mario Fritz, Trevor Darrell, and Pieter Abbeel.
\newblock Parametrized shape models for clothing.
\newblock In {\em 2011 IEEE International Conference on Robotics and
  Automation}, pages 4861--4868, 2011.

\bibitem{Miller2012AGA}
Stephen Miller, Jur~P. van~den Berg, Mario Fritz, Trevor Darrell, Ken Goldberg,
  and P. Abbeel.
\newblock A geometric approach to robotic laundry folding.
\newblock {\em The International Journal of Robotics Research}, 31:249 -- 267,
  2012.

\bibitem{mo2019partnet}
Kaichun Mo, Shilin Zhu, Angel~X Chang, Li Yi, Subarna Tripathi, Leonidas~J
  Guibas, and Hao Su.
\newblock Partnet: A large-scale benchmark for fine-grained and hierarchical
  part-level 3d object understanding.
\newblock In {\em Proceedings of the IEEE/CVF conference on computer vision and
  pattern recognition}, pages 909--918, 2019.

\bibitem{pymeshlab}
Alessandro Muntoni and Paolo Cignoni.
\newblock {PyMeshLab}, Jan. 2021.

\bibitem{10.1145/2366145.2366171}
Rahul Narain, Armin Samii, and James~F. O'Brien.
\newblock Adaptive anisotropic remeshing for cloth simulation.
\newblock {\em ACM Trans. Graph.}, 31(6), nov 2012.

\bibitem{qi2017pointnet}
Charles~R Qi, Hao Su, Kaichun Mo, and Leonidas~J Guibas.
\newblock Pointnet: Deep learning on point sets for 3d classification and
  segmentation.
\newblock In {\em Proceedings of the IEEE conference on computer vision and
  pattern recognition}, pages 652--660, 2017.

\bibitem{qi2017pointnetplusplus}
Charles~R Qi, Li Yi, Hao Su, and Leonidas~J Guibas.
\newblock Pointnet++: Deep hierarchical feature learning on point sets in a
  metric space.
\newblock {\em arXiv preprint arXiv:1706.02413}, 2017.

\bibitem{qiao2021differentiable}
Yiling Qiao, Junbang Liang, Vladlen Koltun, and Ming Lin.
\newblock Differentiable simulation of soft multi-body systems.
\newblock {\em Advances in Neural Information Processing Systems},
  34:17123--17135, 2021.

\bibitem{qiao2020scalable}
Yi-Ling Qiao, Junbang Liang, Vladlen Koltun, and Ming~C Lin.
\newblock Scalable differentiable physics for learning and control.
\newblock {\em arXiv preprint arXiv:2007.02168}, 2020.

\bibitem{seita2021learning}
Daniel Seita, Pete Florence, Jonathan Tompson, Erwin Coumans, Vikas Sindhwani,
  Ken Goldberg, and Andy Zeng.
\newblock Learning to rearrange deformable cables, fabrics, and bags with
  goal-conditioned transporter networks.
\newblock In {\em 2021 IEEE International Conference on Robotics and Automation
  (ICRA)}, pages 4568--4575. IEEE, 2021.

\bibitem{shi2021skeleton}
Ruoxi Shi, Zhengrong Xue, Yang You, and Cewu Lu.
\newblock Skeleton merger: an unsupervised aligned keypoint detector.
\newblock In {\em Proceedings of the IEEE/CVF Conference on Computer Vision and
  Pattern Recognition}, pages 43--52, 2021.

\bibitem{sundaresan2022diffcloud}
Priya Sundaresan, Rika Antonova, and Jeannette Bohg.
\newblock Diffcloud: Real-to-sim from point clouds with differentiable
  simulation and rendering of deformable objects.
\newblock {\em arXiv preprint arXiv:2204.03139}, 2022.

\bibitem{tiwari20sizer}
Garvita Tiwari, Bharat~Lal Bhatnagar, Tony Tung, and Gerard Pons-Moll.
\newblock Sizer: A dataset and model for parsing 3d clothing and learning size
  sensitive 3d clothing.
\newblock In {\em European Conference on Computer Vision ({ECCV})}. {Springer},
  August 2020.

\bibitem{Wang:2011:DDE}
Huamin Wang, Ravi Ramamoorthi, and James~F. O'Brien.
\newblock Data-driven elastic models for cloth: Modeling and measurement.
\newblock {\em ACM Transactions on Graphics}, 30(4):71:1--11, July 2011.
\newblock Proceedings of ACM SIGGRAPH 2011, Vancouver, BC Canada.

\bibitem{xiang2020sapien}
Fanbo Xiang, Yuzhe Qin, Kaichun Mo, Yikuan Xia, Hao Zhu, Fangchen Liu, Minghua
  Liu, Hanxiao Jiang, Yifu Yuan, He Wang, et~al.
\newblock Sapien: A simulated part-based interactive environment.
\newblock In {\em Proceedings of the IEEE/CVF Conference on Computer Vision and
  Pattern Recognition}, pages 11097--11107, 2020.

\bibitem{xue2022useek}
Zhengrong Xue, Zhecheng Yuan, Jiashun Wang, Xueqian Wang, Yang Gao, and Huazhe
  Xu.
\newblock Useek: Unsupervised se (3)-equivariant 3d keypoints for generalizable
  manipulation.
\newblock {\em arXiv preprint arXiv:2209.13864}, 2022.

\bibitem{yu2023diffclothai}
Xinyuan Yu, Siheng Zhao, Siyuan Luo, Gang Yang, and Lin Shao.
\newblock Diffclothai: Differentiable cloth simulation with intersection-free
  frictional contact and differentiable two-way coupling with articulated rigid
  bodies.
\newblock In {\em 2023 IEEE/RSJ International Conference on Intelligent Robots
  and Systems (IROS)}. IEEE, 2023.

\bibitem{Zhang_2017_CVPR}
Chao Zhang, Sergi Pujades, Michael~J. Black, and Gerard Pons-Moll.
\newblock Detailed, accurate, human shape estimation from clothed 3d scan
  sequences.
\newblock In {\em The IEEE Conference on Computer Vision and Pattern
  Recognition (CVPR)}, July 2017.

\bibitem{9156843}
Yuwei Zhang, Peng Zhang, Chun Yuan, and Zhi Wang.
\newblock Texture and shape biased two-stream networks for clothing
  classification and attribute recognition.
\newblock In {\em 2020 IEEE/CVF Conference on Computer Vision and Pattern
  Recognition (CVPR)}, pages 13535--13544, 2020.

\bibitem{zheng20223d}
Zhedong Zheng, Jiayin Zhu, Wei Ji, Yi Yang, and Tat-Seng Chua.
\newblock 3d magic mirror: Clothing reconstruction from a single image via a
  causal perspective.
\newblock {\em arXiv preprint arXiv:2204.13096}, 2022.

\bibitem{zhou2016thingi10k}
Qingnan Zhou and Alec Jacobson.
\newblock Thingi10k: A dataset of 10,000 3d-printing models.
\newblock {\em arXiv preprint arXiv:1605.04797}, 2016.

\bibitem{9721534}
Jihong Zhu, Andrea Cherubini, Claire Dune, David Navarro-Alarcon, Farshid
  Alambeigi, Dmitry Berenson, Fanny Ficuciello, Kensuke Harada, Jens Kober,
  Xiang Li, Jia Pan, Wenzhen Yuan, and Michael Gienger.
\newblock Challenges and outlook in robotic manipulation of deformable objects.
\newblock {\em IEEE Robotics \& Automation Magazine}, 29(3):67--77, 2022.

\end{thebibliography}
}

\end{document}